\documentclass[11pt, oneside]{article}   	
\usepackage{geometry}                		
\usepackage{graphicx}				
\usepackage{amssymb}

\usepackage{amstext}

\usepackage{amsmath}

\usepackage{bbm}

\usepackage[table]{xcolor}
\usepackage{subcaption}
\usepackage{wrapfig}

\title{Automatic Streaming Segmentation of Stereo Video Using Bilateral Space}
\author{Wenjing Ke, Yuanjie Zhu, Lei Yu\\Beihang University}

\begin{document}
\maketitle
\begin{abstract}
In the field of video segmentation, the majority methods are based on monocular video. Traditional unsupervised segmentation algorithms do not perform well in terms of time efficiency and accuracy, because of the bottleneck on the foreground definition. Semi-supervised segmentation algorithms aim to propagate the label information in one or more key frames, which are generated manually and used as masks in the processing, to the whole video. They can achieve high accuracy, while they are not suitable for the application scenario without human interaction. In this paper, we take advantage of binocular camera and propose an unsupervised algorithm to efficiently extract foreground part from stereo video. The depth information is embedded into a bilateral grid in the graph cut model which achieves considerable segmenting accuracy without human interaction. Streaming processing model is integrated to enable on-line processing for stereo video with arbitrary length. The precision, time efficiency, and adaptation to complex natural scenario of our algorithm are evaluated by experiments comparing with state-of-the-art algorithms in both unsupervised and semi-supervised approaches.
\end{abstract}

\section{Introduction}

 Video foreground segmentation is to partition the video image into foreground and background regions with spatial and temporal consistency. It is a fundamental problem in computer vision and has numerous applications in video surveillance, video editing, object tracking and object detection.

Although many image segmentation techniques have made great progress and achieved good segmentation effect, video segmentation is still a challenging problem due to its computational complexity and visual ambiguities. For monocular video, algorithms are normally difficult to define the region of foreground by only color and motion information without human interaction. Therefore, most researches \cite{Choi12, Li13, Marki16, Caelles16} have adapted the approach of providing manually the mask of key frames to facilitate segmentation. Based on this approach, video segmentation systems \cite{Wang05, Bai09, Price09, Fan15} which require gradually adding the user's input to correct the result during segmentation processing are built, and they could achieve considerable segmenting accuracy under human interaction. Considering the scenarios where no input frames are provided, the segmentation algorithm needs additional information which is provided by special equipments, such as flash lights \cite{Sun06}, camera sequences \cite{Joshi06}, RGBD cameras \cite{Hickson14, Zhao15} and time-of-flight cameras \cite{Wang10} in the purpose of achieving good segmenting contours. However, these devices have their own limitations, and are expensive which makes the relevant algorithm hardly to be put into use. Therefore, achieving unsupervised and high accuracy segmentation with portable and non expensive devices is still a challenging work.

In recent years, with continuous progress of stereo matching algorithms in time efficiency and spatial efficiency, binocular cameras \cite{Derek10} which are composed of two synchronized closely and abreast placed cameras to simulate human eyes have attracted attention. Similar to RGBD camera, binocular camera can not only restore three-dimensional space information to some extent, but has the advantage of lower cost and being able to work under natural light. Besides, binocular device can be easily equipped with computer and has already adopted to some mobile phones, so it has a far-ranging applied space and applied outlook. In this paper, we make use of binocular camera and takes depth information as prior information for defining the foreground part.

Besides, the segmentation is performed in the bilateral space. We embed the image into bilateral grid, which is a data structure presented in \cite{Chen07} as an efficient method to accelerate edge-aware processing. Paper \cite{Marki16} is the first to use bilateral grid in semi-supervised video segmentation to reduce the amount of vertex in graph cut which increases greatly its time efficiency. However, there is several limitations in the work \cite{Marki16}. Firstly, it builds a large graph for the entire video and processes the whole graph by one time, so it couldn't process video of long length. Secondly, it's limited as semi-supervised video segmentation and requires manual mask of key frames as input. In our paper, we largely extend the work of \cite{Marki16} and propose an unsupervised video foreground segmentation system with binocular camera. Our system doesn't need any manual input and can process stereo video of arbitrary length.

In this paper, we mainly make the following contributions: Firstly, we use the image disparity to realize the auto-detection of rough foreground part and largely reduce the amount of calculation. And as we know, we are the first to take advantage of the disparity map as the mask of segmentation and realize unsupervised foreground segmentation.  Secondly, we embed the video frames into bilateral grid to perform an automatic segmentation and provide a guideline on how to manipulate on the bilateral grid, which could be widely transformed to many other applications or algorithms. Thirdly, by integrating the streaming video processing skill, we realize the auto-processing of arbitrary length video.

\section{Related Work}

According to the type of video capture device, video segmentation system can be generally divided into two categories: segmentation system with monocular camera and segmentation system with special device.
\subsection{Segmentation system with monocular camera}

Most researchers deal with the whole video at one time. Graph cut \cite{Rother04} is core technique in this approach. The image data is treated as a vertex in a graph, and the segmentation problem is regarded as a binary labeling problem, which can be solved by an energy optimization processing. Since the efficiency of graph cut depends on the number of graph nodes, it's nearly impossible to use it directly on the whole video. To reduce the number of nodes, many researchers use clustering algorithms such as mean-shift \cite{Paris08}, spectral clustering \cite{Levin08}, or super pixels \cite{Galasso12}. In graph based method, optical flow \cite{Zach07} and nearest neighbor fields \cite{Chen13} are used also by many scholars to build correspondence between adjacent frames.
\subsection{Segmentation system with special device}

Since only two-dimensional information available for a monocular video makes the segmentation challenging, some researchers use other devices to get additional information. Flash matting \cite{Sun06} uses flashlight and argues that foreground lies where the biggest difference happens between a flash image and a no flash image. Bayesian matting algorithm is mainly used in flash matting. Paper \cite{Joshi06} proposed a foreground segmentation system using a series of parallel synchronized cameras. Since the foreground is separate from the background, those cameras can be used to collect foreground objects at different parts of the background. RGBD camera is used in \cite{Hickson14} to provide the depth information   to compute the optical flow and the depth information incorporated into the graph model. In \cite{Zhao15}, the depth map is incorporated into closed-form matting, taking into account the continuity of time and good results are achieved. Time-of-flight camera determines the depth at each pixel by measuring the time it takes for infrared light to travel to the object and return to the camera. In \cite{Wang10}, the depth information provided by the TOF camera is also incorporated into the graph model to achieve real-time segmentation. Binocular device is another trade-off device for RGBD camera. In \cite{Wei09}, the depth map is pre-segmented, and a change detection is performed based on the adjacent frame difference. The foreground is finally obtained by combining it with the edge detection. In \cite{Zhao14}, color, depth and motion information are united to group firstly pixels into spatial and temporal consistent super pixels, and then build a new graph cut model to benefit those super pixels. In \cite{Kolmogorov05}, an energy optimization problem is constructed by fusing likelihoods for stereo matching, color and contrast. The stereo video is efficiently segmented frame-by-frame.

\section{Method}

Let $ \{ \mathcal{V}_l,\mathcal{V}_r \} : \Omega \longrightarrow \mathbb{R}^3$ denote an stereo video. With on user input given, we seek a binary mask $\mathcal{M}:  \Omega \longrightarrow \{0, 1\}$ that labels each pixel in the left video either as foreground or background.

Our approach starts from stereo video stream and it mainly contains three parts: pre-processing(\ref{preprocessing}), graph cut in bilateral space(\ref{bilateral grid}, \ref{graph cut}) and streaming processing(\ref{streaming processing}).

\subsection{Pre-processing}\label{preprocessing}

In this part, we need to find out the valid area where the foreground object may lie in. Suppose that intrinsic and extrinsic parameters of stereo camera parameters are known after a calibration process. We use a traditional algorithm \cite{Fusiello00} to rectify images from two camera views. Then we calculate the disparity map through Semi-global Block Matching algorithm(SGBM) \cite{Hirschmuller08}.

Before applying the segmentation algorithm, we want to find out the valid area where the foreground object may lie in and extract this effective foreground area, thus reducing computational work of subsequent process. In our system, as we suppose that the foreground object is the part closer to the camera in the scene, the ROI should the area with large parallax.

\begin{figure}
    \centering
    \begin{subfigure}[b]{0.22\textwidth}
        \includegraphics[width=\textwidth]{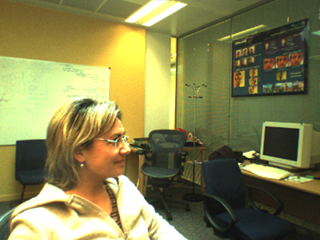}
        \caption{Origin image}
        \label{fig: origin}
    \end{subfigure}
    ~ 
    \begin{subfigure}[b]{0.22\textwidth}
        \includegraphics[width=\textwidth]{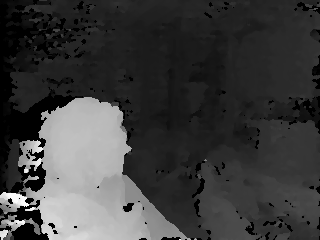}
        \caption{Disparity image}
        \label{fig:disp}
    \end{subfigure}
    \begin{subfigure}[b]{0.28\textwidth}
        \includegraphics[width=\textwidth]{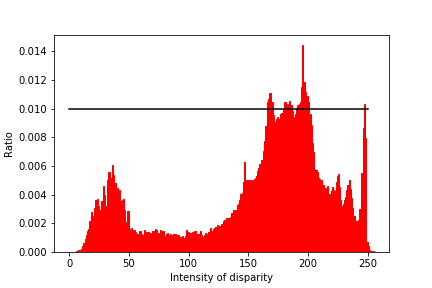}
        \caption{Histogram}
        \label{fig:hist}
    \end{subfigure}
    ~ 
    \begin{subfigure}[b]{0.22\textwidth}
        \includegraphics[width=\textwidth]{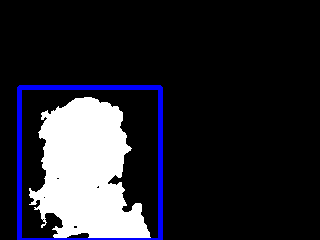}
        \caption{Region of interest}
        \label{fig:contour}
    \end{subfigure}
    \caption{Pictures obtained in different steps}\label{fig1}
\end{figure}

For an object in the scene, the closer it is to a camera, the larger its disparity is. Thus the problem of finding the foreground part equals to find the part with larger disparity. By counting up the histogram of the disparity image as figure \ref{fig:hist}, we may firstly calculate the disparity threshold $d_{th}$ of foreground part by comparing frequency of disparity:
\begin{equation}
\begin{cases}
f(d)>f(d-1)  \&   f(d)>f(d+1),  & \cr
f(d)>n_{th1} & \cr
\end{cases}
\end{equation}
where $f(d)$ is the frequency of disparity $d$,  $n_{th1} = \frac{\text{Total amount of pixels}}{100}$ and  the denominator of 100 is optimal empirically selected.  With respect to this, we obtain a depth threshold $d_{th}$ with max frequency of disparity.

Then we expand $d_{th}$ to an interval $\Gamma_{d} = [d_{th}-\Delta d_1, d_{th}+\Delta d_2]$ by applying region-growing algorithm on the disparity image with respect to:
\begin{equation}
\forall d \in  \Gamma_{d_{th}}, \sum_{d_0 \in \Gamma_d}f(d_0)>n_{th2}
\end{equation}
where $n_{th2} = \frac{\text{Total amount of pixels}}{10}$, it means that we initialize the foreground part as $\frac{1}{10}$ of whole image. The pixels falling in $\Gamma_{d}$ are the initial disparity mask and the minimum bounding rectangle as show in figure \ref{fig:contour} of the disparity mask is regarded as the valid region of interest.


\subsection{Bilateral grid}\label{bilateral grid}

Bilateral grid \cite{Chen07} has recently been introduced as a data structure that enables fast edge-aware image processing. Bilateral grid is to lift an image into a multi-dimensional space. Both the coordinate (x, y) and the three color channels (Y, U, V) of image are also extended as spatial dimension, intensity dimension and chroma dimension. The image pixels are converted to many points distributing in the bilateral grid.

As lifting the pixels into the grid, the grid size of each dimension is down sampled and it's much lower than image lattice. In bilateral grid, a grid could be encompassed by several pixels in all dimensions, thus operations in bilateral grid can be more efficient since number of grids is much fewer than pixels. A traditional bilateral grid is a five-dimensional array , including spatial information, image intensity and chroma. In our work, we take the idea in \cite{Marki16} and likely express input videos as a higher-dimensional bilateral space, which concatenate pixel intensity and chroma information(y, u, v), spatial information(x, y, depth) and temporal information(t), where the depth information is represented by a disparity mask gained in section \ref{preprocessing}.

\paragraph{Grid creation}
Given an image sequence $P=\{p | p=[x, y, t]\}$ and the corresponding disparity sequence $D = \{d|d = d(x,y,t)\}$. The first step is lifting pixels into bilateral space by down sampling. Let $B$ be the bilateral space and the dimension of our bilateral grid $\Gamma$ be $l_Y\times l_U\times l_V\times l_x\times l_y\times l_d\times l_t$ . The sampling rate on each dimension is computed as
\begin{equation}
\gamma_i=\frac{l_i}{i_{max}-i_{min}},i=Y,U,V,x,y,d,t
\end{equation}
This allows us to create a mapping from pixel p to the bilateral space
\begin{equation}
\begin{gathered}
P \longrightarrow B\\
b(p)=[\gamma_YY,\gamma_UU,\gamma_VV,\gamma_xx,\gamma_yy,\gamma_dd,\gamma_tt]
\end{gathered}
\end{equation}
So we can express our bilateral data as a $P\times7$ matrix. The pixels are all embedded into the grid and $b(p)$ has influence on all dimensions of grid vertex it situated in. To compute the influence of $b(p)$  on vertex $v$, we use the adjacent weighting method mentioned in \cite{Marki16} by representing influence with a weight value and supposing $b(p)$ only has influence on the nearest vertex and its neighbors in each dimension. The weight function $\omega(v,b(p))$ determines the weight of $b(p)$  on vertex v. So, the closer $b(p)$ and $v$ are, the greater  $\omega(v,b(p))$ becomes. Note $N_{b(p)}$ as the nearest vertex of $b(p)$, $\omega(v,b(p))$ is computed as product of distance in each dimension as follows.
\begin{equation}\label{Weight}
\omega(v,b(p))=
\begin{cases}
\prod(1-|v_i-b(p)_i|), &if \sum|v_i-N_{b(p)_i}| \leq 1\cr
0, & else
\end{cases}
\end{equation}
where $v_i$, $b(p)_i$ and $N_{b(p)_i}$ denotes the i-th coordinates of $v$, $b(p)$ and $Nb(p)$ separately. The total weight $S(v)$ on a vertex is obtained simply by accumulating  $\omega(v,b(p))$ for all possible pixels. So if there exists more pixels corresponding to some vertex, it will have a larger weight. Through this process of grid creation, we are capable of expressing the distribution of space, time and color information in bilateral space in form of vertex weight.
\begin{equation} S(v)=\sum\limits_{p\in P}\omega(v,b(p)) \end{equation}
Through this process of grid creation, we are capable of expressing the distribution of space, time and color information in bilateral space in form of vertex weight.

\subsection{Graph cut}\label{graph cut}

In this step, we try to solve the label $\alpha$ for each vertex in the bilateral grid, that $\alpha=0$ marks background and $\alpha=1$ marks foreground.

We use a generally used graph cut method. By representing the video as a graph $\mathcal{G} = <\mathcal{T}, \mathcal{E}>$, where $\mathcal{T}$ is the set of grid vertices and $\mathcal{E}$ is the set of edges that connect adjacent points of the grid, the issue of segmentation is turned to the global optimization of energy equation:
\begin{equation}
\begin{gathered}
\alpha=\mathop{\arg\min}\limits_{\alpha=\{\alpha_v|\alpha=0or1\}}E(\alpha)\\
E(\alpha)=\lambda\sum\limits_{v\in\mathcal{T}}\theta_v(v,\alpha_v)+\sum\limits_{(u,v)\in\mathcal{E}}\theta_{uv}(u,\alpha_u,v,\alpha_v)
\end{gathered}
\end{equation}

The total energy $E(\alpha)$ includes data term $\theta_v$ and pairwise term $\theta_{uv}$ balanced by a constant coefficient $\lambda$. The data term stands for the cost of labeling a pixel to fg/bg, the pairwise term stands for the cost of assigning same fg/bg label to the adjacent pixels in the same frame. This form of energy can be minimized by a standard max-flow/min-cut algorithm \cite{Rother04}. Since the total number of vertices is far less than that of image pixels, the  amount of calculation is reduced.

The data term $\theta_v (v, \alpha_v)$ represents the cost of each vertex belonging to the label $\alpha_v$. It depends on mask which retains the coherent vertex data for each label term as:
\begin{equation}
\begin{gathered}
S_{BG}(v)=\sum\limits_{p\in\mathcal{P}}\omega(v,b(p))\mathbb{I}_{BG}(p)\\
S_{FG}(v)=\sum\limits_{p\in\mathcal{P}}\omega(v,b(p))\mathbb{I}_{FG}(p)
\end{gathered}
\end{equation}
where $\mathbb{I}_{\times}(p)$ indicates mask information and it equals to 1 if  $p \in \times$ and 0 otherwise. Since we use disparity information instead of manual mask, we multiply the weights to $\frac{b(p)_d}{l_d}$ which ranges in $[0,1]$. As the disparity is small, the cost of belonging to background gets small and the cost of belonging foreground gets large. The formula turns into:
\begin{equation}\label{DispSplatted}
\begin{gathered}
S_{BG}(v)=\sum\limits_{p\in\mathcal{P}}\omega(v,b(p))\frac{b(p)_d}{l_d}\\
S_{FG}(v)=\sum\limits_{p\in\mathcal{P}}\omega(v,b(p))(1-\frac{b(p)_d}{l_d})
\end{gathered}
\end{equation}
And we have thus the data term.
\begin{equation}\label{DispDataTerm}
\theta_v(v, \alpha_v)=
\begin{cases}
S_{BG}(v), & \alpha_v=1 \cr
S_{FG}(v), & \alpha_v=0 \cr
\end{cases}
\end{equation}

The pairwise term represents the cost of assigning same/different label to adjacent vertex and it ensures that adjacent vertices tend to have similar labels. Taking advantage of the vector and multi-dimension representation of vertex, the pairwise term can be computed as:
\begin{equation}
\theta_{uv}(u, \alpha_u, v, \alpha_v)=
\begin{cases}
g(u,v)S(u)S(v), & if \alpha_u\neq\alpha_v \cr
0, & else \cr
\end{cases}
\end{equation}
where  $g(u,v)$ is a 7-dimensional Gaussian kernel. The coefficient $\lambda$ balances data term and smooth term. Since disparity information is not accurate, we set $\lambda$ to be around 1 to balance the accuracy of total energy.

\subsection{Streaming processing}  \label{streaming processing}

Streaming processing \cite{Xu12} is a video processing method which enables operation for video of arbitrary length in an on-line streaming way. In our binary segmentation problem, we divide the video into $m$ sub-sequences $\mathcal{V}=\{V_1, V_2, ...V_m\}$ and the corresponding segmentation is $\mathcal{S}=\{S_1,S_2,...S_m\}$. We assume that the segmentation result of sequence $V_i$ is only related to the previous subsequence $V_{i-1}$ and its corresponding segmentation result $S_{i-1}$. The streaming segmentation of a video is thus modeled as:
\begin{equation}
\begin{aligned}
\mathcal{S}=\{&S_1,S_2,...S_m\}\\
=\bigg\{&\arg\min\limits_{S_1}E(S1|V1), ...\arg\min\limits_{S_i}E(S_i|V_i, S_{i-1}, V_{i-1}), ...\\    &\arg\min\limits_{S_m}E(S_m|V_m, S_{m-1}, V_{m-1})\bigg\}
\end{aligned}
\end{equation}
where $E(S1|V1)$ is the segmentation energy of the first subsequence and $E(S_m|V_m, S_{m-1}, V_{m-1})$ corresponds to energy of current subsequence given previous segmentation result.

Based on that streaming processing model, we involve our disparity based bilateral space graph cut to construct a streaming video segmentation system. Since the first subsequence has no previous information in time, we must use our disparity information as prior to obtain $E(S_1|V_1)$. The automatic bilateral space graph cut model using disparity as prior information is used to represent $E(S_1|V_1)$, that is
\begin{equation}
E(S_1|V_1)=\lambda\sum\limits_{v\in\mathcal{T}}\theta_v(v,\alpha_v)+\sum\limits_{(u,v)\in\mathcal{E}}\theta_{uv}(u,\alpha_u,v,\alpha_v)
\end{equation}
The data term is estimated using disparity weights of bilateral vertex. $E(S_i|V_i, S_{i-1}, V_{i-1})$ describe the influence of previous sequence on current sequence. In this case, both segmentation result of previous sequence and disparity prior of current sequence contribute to the data term.
\begin{equation}
\begin{aligned}
E(S_i|V_i, S_{i-1}, V_{i-1})=&\sum\limits_{v\in\mathcal{T}}(\lambda_i\theta_v^i(v,\alpha_v)+\lambda_d\theta_v^d(v,\alpha_v))\\
+&\sum\limits_{(u,v)\in\mathcal{E}}\theta_{uv}(u,\alpha_u,v,\alpha_v)
\end{aligned}
\end{equation}
where two constant coefficients $\lambda_i$ and $\lambda_d$ are set to balance those three terms.  $\theta_v^d(v,\alpha_v)$ is computed using (\ref{DispSplatted}) and (\ref{DispDataTerm}).  $\theta_v^i(v,\alpha_v)$ is computed with vertex weights of previous segmentation mask as follows.
\begin{equation}
\theta_v^i(v, \alpha_v)=
\begin{cases}
S_{BG}^i(v)=\sum\limits_{p\in P_{i}}\omega(v,b(p))\mathbb{I}_{BG}(p), & \alpha_v=0 \cr
S_{FG}^i(v)=\sum\limits_{p\in P_{i}}\omega(v,b(p))\mathbb{I}_{FG}(p), & \alpha_v=1 \cr
\end{cases}
\end{equation}
where $\omega(v,b(p))$ is the same as (\ref{Weight}) and $\mathbb{I}_{BG}(p)$, $\mathbb{I}_{FG}(p)$ are two indicators of previous segmentation mask.

\section{Experiments and Results}
\begin{figure}
\centering
	\begin{subfigure}[b]{\textwidth}
	\includegraphics[width=0.24\textwidth]{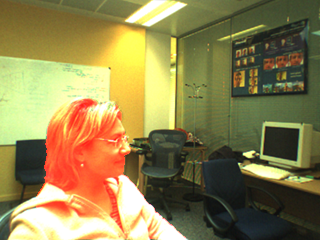}
	\includegraphics[width=0.24\textwidth]{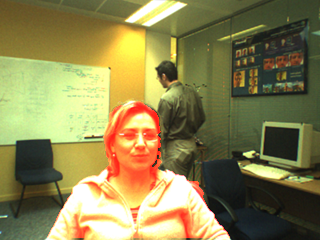}
	\includegraphics[width=0.24\textwidth]{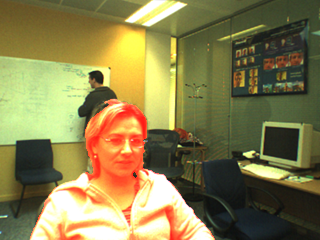}
	\includegraphics[width=0.24\textwidth]{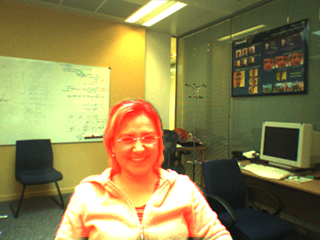}
	\end{subfigure}
	
	\begin{subfigure}[b]{\textwidth}
	\includegraphics[width=0.24\textwidth]{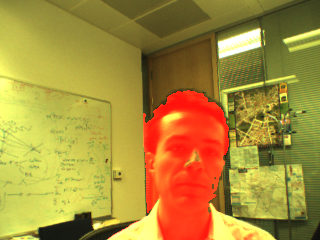}
	\includegraphics[width=0.24\textwidth]{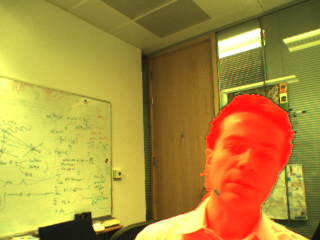}
	\includegraphics[width=0.24\textwidth]{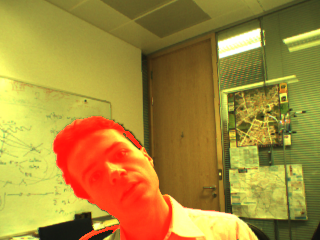}
	\includegraphics[width=0.24\textwidth]{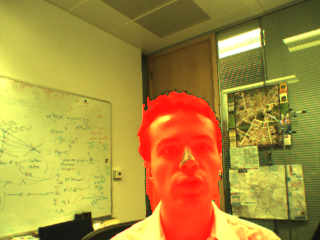}
	\end{subfigure}
	
\caption{Results of our approach on i2iDatabase: video IU(a) and video AC(b)} \label{resultsIU_AC}
\end{figure}
 
Evaluation of our method is difficult because almost all public benchmarks provide only monocular video data in the field of video segmentation. The experimental dataset i2iDatabase  is presented and used in \cite{Kolmogorov05} to evaluate their binocular video foreground extraction algorithms. The video data set consists of a group of video captured with a binocular camera device, including both static background and dynamic background videos. Some of the videos in the data set contain the segmentation ground-truth that is manually marked for every 5 frames. In ground-truth, each pixel is marked as background and foreground. We use this stereo video dataset to evaluate our segmentation system both in terms of accuracy and efficiency. Our approach is implemented in C++. The system runs on a desktop with a 3.4 GHz 6-Core Intel i5-4670 CPU and 8 GB RAM.

\paragraph{Metrics}
DAVIS \cite{Perazzi16} is a recently published benchmark to evaluate techniques in the domain of video segmentation, but unfortunately it doesn't contain any stereo video set. So we use 5 video set AC, IU, JM, MS, VK in i2iDatabase dataset (results shown in Figure \ref{resultsIU_AC}) and three measures used in DAVIS database to evaluate our system in the following terms: region similarity $\mathcal{J}$ (with respect to intersection of union - IoU), contour accuracy $\mathcal{F}$ and temporal stability $\mathcal{T}$. Although our method focus on the case with no manual input, we compare our result with the state-of-the-art methods in both unsupervised (FST \cite{Papazoglou13}) and semi-supervised techniques (BVS \cite{Marki16} and OSVOS \cite{Caelles16}), the latter of which takes ground-truth of first frame as initial mask.

\begin{table*}
\centering
\begin{tabular}{c|ccccc}
\hline
   \bf Measure &\bf Ours($l = 3$) &\bf Ours($l = 10$) &\bf OSVOS &\bf BVS  &\bf FST\\
\hline
\bf $\mathcal{J}$  \bf Mean $\uparrow$ &\ 89.0 & 85.7 &\bf\ 94.6  &\ 82.6 &\ 43.1 \\
\hline
 \bf $\mathcal{J}$ \bf Recall $\uparrow$ &\bf\ 99.1 &\ 98.9 & 98.8 &\ 96.6 &\ 42.3 \\
\hline
\bf $\mathcal{J}$ \bf Decay $\downarrow$ &\ -2.5 &\bf\ -4.5 & 0.8 &\ 0.7 &\ 2.1 \\
\hline
\bf $\mathcal{F}$ \bf Mean $\uparrow$ &\ 69.7 & 63.4 &\bf\ 87.0  &\ 48.7 &\ 36.0 \\
\hline
\bf $\mathcal{F}$ \bf Recall $\uparrow$ &\ 94.4 &\ 81.8 &\bf\ 99.5 &\ 54.1 &\ 15.3 \\
\hline
\bf $\mathcal{F}$ \bf Decay $\downarrow$ &\ -1.9 &\bf\ -3.4 &3.8 &\ 3.0 &\ 2.0 \\
\hline
\bf $\mathcal{T}$ $\downarrow$ &\ 49.0 &\ 47.2 &\bf\ 46.8 &\ 49.8 &\ 54.4 \\
\hline
\end{tabular}
\caption{Comparison of our approach to the state of the art on DAVIS}\label{tab:Table1}
\end{table*}

\begin{table*}\label{Table2}
\centering
\begin{tabular}{c|ccccc}
\hline
 \bf Sequence &\bf Ours(\bf $l = 3$) &\bf Ours(\bf $l = 10$) & \bf OSVOS & \bf BVS & \bf FST\\
\hline
 \bf AC  &\ 84.5 & 82.1 &\bf\ 93.8  &\ 76.9 &\ 61.9 \\
\hline
 \bf IU &\bf\ 93.4 &\ 91.3 & 93.1 &\ 85.4 &\ 33.7 \\
\hline
 \bf JM &\ 92.4 &\ 91.2 &\bf\ 95.7 &\ 81.0 &\ 14.9 \\
\hline
 \bf MS &\ 86.4 & 81.8 &\bf\ 94.4  &\ 84.1 &\ 60.8 \\
\hline
  \bf VK &\ 88.0 &\ 82.1 &\bf\ 95.9 &\ 85.5 &\ 44.4 \\
\hline
 \bf Mean &\ 89.0 &\ 85.7 &\bf\ 94.6 &\ 82.6 &\ 43.1 \\
\hline
\end{tabular}
\caption{Per-sequence results of region similarity}\label{tab:Table2}
\end{table*}

\paragraph{Accuracy}
Our algorithm is affected by the length of sub-sequence $l$: larger length means better time consistency but lower accuracy. We have evaluated our algorithm with $l = 3$ and $l = 10$. Table \ref{tab:Table1} shows the overall evaluation of our approach and the state-of-the-arts. Table \ref{tab:Table2} shows the per sequence result of region similarity compared with other algorithms.

We see that OSVOS has the best performance overall, but the advantage of our approach lies in automation and it outperforms also FST and BVS in terms of region similarity, contour accuracy and temporal stability. Our mean performance and contour accuracy are only worse than OSVOS, but recall outperforms the other methods. Also the accuracy of all the other methods show a decay over time, while our algorithm doesn't show such a loss over time.

\begin{figure}
    \centering
    \begin{subfigure}[b]{0.19\textwidth}
        \includegraphics[width=\textwidth]{{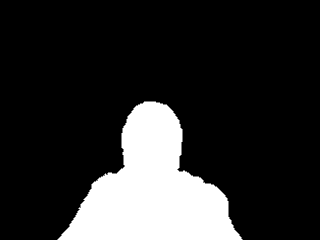}}
        
        \includegraphics[width=\textwidth]{{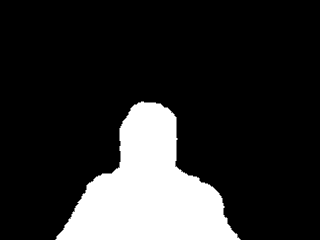}}
        
        \includegraphics[width=\textwidth]{{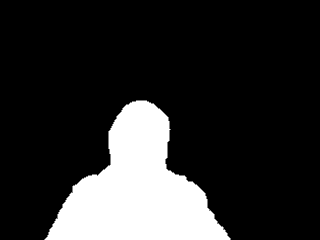}}
        \caption{}
        \label{fig: gt}
    \end{subfigure}
    \begin{subfigure}[b]{0.19\textwidth}
       \includegraphics[width=\textwidth]{{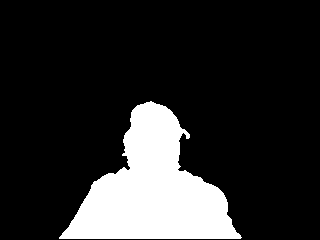}}
       
        \includegraphics[width=\textwidth]{{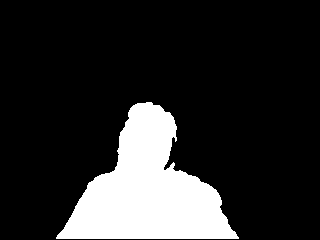}}
        
        \includegraphics[width=\textwidth]{{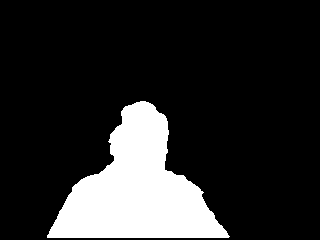}}
        \caption{}
        \label{fig: dbvs}
    \end{subfigure}
    \begin{subfigure}[b]{0.19\textwidth}
       \includegraphics[width=\textwidth]{{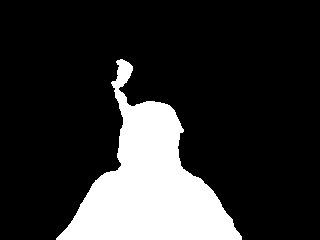}}
       
        \includegraphics[width=\textwidth]{{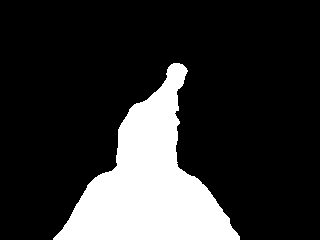}}
        
        \includegraphics[width=\textwidth]{{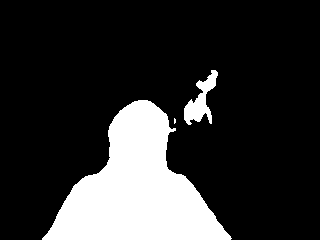}}
        \caption{}
        \label{fig: osvos}
    \end{subfigure}
    \begin{subfigure}[b]{0.19\textwidth}
       \includegraphics[width=\textwidth]{{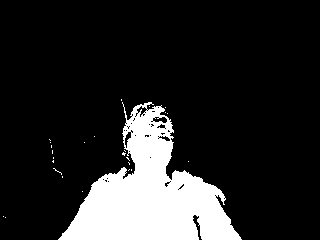}}
       
        \includegraphics[width=\textwidth]{{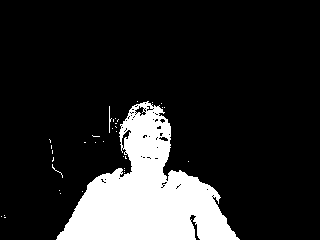}}
        
        \includegraphics[width=\textwidth]{{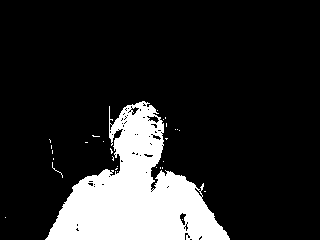}}
        \caption{}
        \label{fig: bvs}
    \end{subfigure}
    \begin{subfigure}[b]{0.19\textwidth}
       \includegraphics[width=\textwidth]{{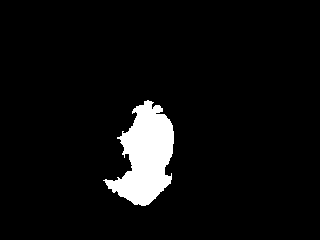}}
       
        \includegraphics[width=\textwidth]{{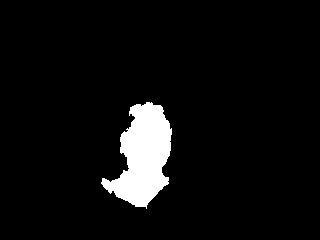}}
        
        \includegraphics[width=\textwidth]{{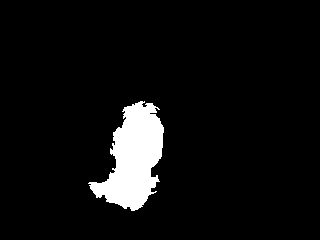}}
        \caption{}
        \label{fig: fst}
    \end{subfigure}
    \caption{Comparison of output on video IU obtained via different algorithm: (a) ground truth, (b) our approach, (c) OSVOS, (c) BVS and (e) FST.} 
    \label{fig: comparison}
\end{figure}

A visualization of results from different algorithms is shown in figure \ref{fig: comparison}. For OSVOS and BVS, we provide the first frame as the mask and they track the target object in the following frames. OSVOS is based on deep learning trained model and it recognizes the target object in the frame. But the wrong classification of some unrelated object results in the decrease of segmentation accuracy. BVS uses bilateral grid to preserve good edge for the object, but this is also its shortcoming because it focuses on the edge and fails to merge the whole object. FST is based on optimal flow and it only works on the moving part. So the result of FST is not good. Reversely, our algorithm works well to segment the foreground part and our result has a high accuracy comparing with the ground truth.

\paragraph{Time}
The use of bilateral grid greatly simplifies the computations of bilateral filter, Gaussian kernel and the pairwise term in Gibbs energy. Previous works \cite{Chen07} \cite{Marki16} have well demonstrated the outstanding time efficiency of bilateral grid and Paper \cite{Marki16} has compared its time efficiency with other segmentation algorithms. For a typical video with 400 $\times$ 600 color images, if we set $l=10$, and [Intensity grid size, Chroma grid size, Spatial grid size, Temporal grid size, Disparity grid size] = [7, 9, 13, 2, 2], the average time to compute the segmentation is $0.87s$, including $0.8s$ for grid converting process and $0.07s$ for graph cut. We have also optimized our program by CUDA, and the time for grid transforming reduced to $0.12s$.

\begin{table}\label{Table3}
\centering
\begin{tabular}{c|l|l|l}
\hline
  \bf   &\bf Sequence  &\bf Grid conversion  & \bf Graph cut \\
\hline
\bf CPU  &\  $l=3$   &\  1.23 s &\  0.19 s \\

              &\  $l=10$ &\  0.8 s    &\  0.07 s  \\
\hline
\bf GPU &\  $l=3$   &\ 0.19 s   &\ -      \\
\bf         &\ $l=10$  &\  0.12 s  &\ -      \\
\hline
\end{tabular}
\caption{Average time cost for a 400 $\times$ 600 video frame}\label{tab:Table3}
\end{table}

\paragraph{Limitation}
Our algorithm has its limitations in two ways: firstly, the segmentation is based on the extracted disparity map, therefore the accuracy of disparity map is the bottleneck in our algorithm. When applied to the scene with poor illumination condition or the object that has few texture, the depth is hard to estimate and it gives poor prior information for following segmentation. Secondly, our algorithm works well on the scene with definite foreground objects but is not applicable for complex scene.

\section{Conclusions}

In this paper, we propose an unsupervised method to accomplish stereo video foreground segmentation. We introduce the creativity of our algorithm on using bilateral grid and taking advantage of scene disparity. By experiment, we test the accuracy and practicability on i2iDatabase and also compared our algorithm with the state-of-art. Experiments show that we outperform not only the state of art in terms of unsupervised techniques, and even some semi-supervised algorithms like BVS. In addition, although there is one point gap between OSVOS and our algorithm, the ours have the advantage of no user input and thus lager application prospect.

Some future work could be done to extend our work. In our experiments, the calculation of disparity adopts the classical SGBM algorithm, which is much inferior to the state of art. The future research could inherit current work and attempt to merge Stereo matching and bilateral grid into the graph cut model with global optimization after that.

\end{document}